\documentclass{article}

\usepackage[nonatbib,preprint]{neurips_2026}
\usepackage[square, comma, sort&compress, numbers]{natbib}
\usepackage[utf8]{inputenc} 
\usepackage[T1]{fontenc}    
\usepackage{url}            
\usepackage{booktabs}       
\usepackage{todonotes}
\usepackage{amsfonts}       
\usepackage{amsmath}        
\usepackage{amssymb}        
\usepackage{nicefrac}       
\usepackage{microtype}      
\usepackage[dvipsnames]{xcolor}         
\usepackage{graphicx}       
\usepackage[smallerops]{newtxmath}

\usepackage{hyperref}
\hypersetup{
    colorlinks=true,
    linkcolor=NavyBlue,
    citecolor=NavyBlue,
    urlcolor=NavyBlue
}

\newcommand{\R}{\mathbb{R}}
\DeclareMathOperator{\tr}{Tr}

\DeclareMathOperator{\diag}{diag}
\DeclareMathOperator*{\EE}{\mathbf{E}}

\title{Fisher information and the geometry of memorization in neural networks}

\author{%
  Daniel T.~Bernstein\textsuperscript{*}\\
  Princeton University\textsuperscript{1}
  \And
  Chase Goddard\textsuperscript{*}\\
  Princeton University\textsuperscript{2}
  \And
  Luca Di Carlo\\
  Princeton University\textsuperscript{1,2}
  \And
  Vudtiwat Ngampruetikorn\textsuperscript{\S}\\
  University of Sydney\textsuperscript{3}
  \And
  David J.~Schwab\textsuperscript{\S}\\
  CUNY Graduate Center\textsuperscript{4}
}

\begin{document}
\maketitle
\def\thefootnote{*}
\footnotetext{DTB and CG contributed to this work equally.}
\def\thefootnote{\S}
\footnotetext{VN and DJS contributed to this work equally.}
\begingroup
\renewcommand{\thefootnote}{1}
\footnotetext{Lewis-Sigler Institute\quad
\textsuperscript{2} Joseph Henry Laboratories\quad
\textsuperscript{3} School of Physics\quad
\textsuperscript{4} Initiative for the Theoretical Sciences}
\endgroup
\def\thefootnote{\arabic{footnote}}
\begin{abstract}
A fruitful approach towards understanding generalization in machine learning involves characterizing how models separate generalizable structure from idiosyncratic or even corrupted features of training data.
Fisher information has emerged as a diagnostic tool for this separation, leveraging loss geometry to distinguish generalization from memorization. 
Here we investigate this connection in vision models trained on CIFAR-10 with controlled noise that can be learned only through memorization.
We show that projecting model weights onto the top Fisher eigenvectors decouples general task performance from noisy sample memorization, and we investigate how this decoupling presents differentially across layers. 
We show that the same phenomena are present in small transformers trained to solve a synthetic in-context learning task, and are reproduced in closed form in a simple, analytically solvable linear model.
Our results reveal that Fisher information captures aspects of the geometric structure underlying how neural networks allocate capacity between generalization and memorization.
\end{abstract}

\section{Introduction}

Modern neural networks can memorize their training set~\cite{zhang2017understanding}, yet in practice they learn features that generalize to unseen examples. Disentangling memorization from generalization is a central problem in machine learning---large language models reproduce substantial fractions of their pretraining data~\cite{carlini2021extracting}, raising copyright and privacy concerns; adversaries can exploit memorization through data poisoning~\cite{goldblum2020dataset}; and memorization of mislabeled examples degrades generalization. Progress on any of these fronts requires a clearer picture of how trained models store memorized content and what distinguishes it from generalizing features. 

The geometry of the loss landscape offers a natural setting for this question, and recent work argues that the eigenspectrum of the Fisher information separates memorized information from generalizing features, using this separation to remove memorized content from trained language models~\cite{merullo2025memorization}. Here, we investigate this phenomenon across several settings designed to probe how and when the Fisher encodes memorized features.

The Fisher information's averaging structure suggests why it might isolate memorization. On a dataset $D$, the Fisher decomposes as an average of per-sample contributions, $F = \frac{1}{|D|}\sum_{x\in D} F_x$, where each $F_x$ measures the sensitivity of the network's output to its parameters at sample $x$. Directions that encode generalizing features should align across single-sample Fishers and add coherently in this average. On the other hand, directions that encode memorization of particular samples should instead be largely orthogonal in parameter space and suppressed by a factor \char`~$1/|D|$ in the average. The top of the population Fisher's eigenspectrum is therefore a natural place to look for a separation between memorized content and generalizing structure.

This picture turns memorization into a set of testable geometric predictions. If memorized features are supported on sample-specific directions, then projecting onto the top Fisher eigenspaces should preserve structure shared across the training distribution while suppressing information tied to individual examples. In a trained network, however, the relevant eigenspaces may depend on the data used to compute the Fisher, may appear only in particular layers, and may trade off memorization removal against the loss of weak but useful features. These considerations motivate our experiments. The Gaussian-image setting isolates pure input-level memorization (\S\ref{sec:gaussimage}), the label-noise setting tests whether de-memorization can recover an underlying correct label (\S\ref{sec:labelflips}) and the in-context regression setting asks whether the same geometry applies to memorized latent tasks (\S\ref{sec:icl}). The barcode model then distills the same mechanism into a linear setting where finite-sample correlations with uninformative features can be analyzed in closed form (\S\ref{sec:barcode}).

Our work extends this Fisher-based account of memorization in three directions. 
First, we show that distinct forms of memorized content---random-input examples, mislabeled training examples, and memorized latent tasks in a synthetic in-context learning task---leave distinct signatures in the layer-wise Fisher geometry.
Second, we find that the geometric separation between memorization and generalization depends sharply on the data used to evaluate the Fisher: a denoised dataset yields cleaner separation than the full training distribution. 
Third, suppressing memorized directions can partially recover task generalization that the trained model otherwise loses to overfitting.

\subsection{Contributions}
In this work, we study the relationship between sample memorization and the Fisher information. We do so by investigating conditions under which the projection of a neural network's weights onto a subspace governed by the Fisher information induces de-memorization. We define de-memorization as an operation performed on a neural network which disproportionately degrades its performance on samples that the network was forced to memorize, while inducing only minor degradation on other samples. We emphasize that de-memorization is separate from the phenomenon of unlearning: de-memorization does not allow a network to selectively forget individual samples in its training set \citep{cao2015towards,ginart2019making,guo2020certified}. We specify the projection operation in detail in the Experiments section. Our contributions are as follows:
\begin{itemize}
    \item We demonstrate how low-eigenvalue subspaces of the Fisher information encode memorized data across a variety of experimental settings: We investigate input/output noise in image classifiers, the structure of subspaces across layers, and an in-context regression task.
    
    \item We show that de-memorization performance depends on the samples used to compute the Fisher: the structure of Fisher eigenspaces with respect to memorization is data-dependent.
    
    \item In an in-context regression problem, we demonstrate that de-memorization can be lifted to the level of latent tasks while simultaneously improving task generalization, and we identify model components responsible for task memorization.

    \item We investigate how easily individual samples can be de-memorized from the model by using information encoded in single-sample Fisher matrices, and we use the language of participation ratios to understand how different kinds of samples contribute to different regions of the eigenspectrum.
    
    \item We develop an analytically tractable ``barcode'' model that exhibits the de-memorization phenomenon in a simple linear setting.
\end{itemize}

\section{Related Work}
\paragraph{Curvature-based signatures of memorization.}%
Recent works have used local curvature measures to identify example-level memorization. Most closely related to our work, \citet{merullo2025memorization} study the eigenspectrum of loss curvature in language models, finding that the top Fisher modes encode reasoning-like structure, whereas lower-curvature modes encode memorized samples. Loss curvature per-sample has been proposed as a way to identify mislabeled and conflicting examples, and corrupted data injected into the training set can exhibit high-curvature signatures \citep{garg2024memorization}. Influence-function methods use gradients and Hessian-vector products to map predictions back to influential training examples \citep{koh2017understanding}. 

\paragraph{Memorization and noisy labels.}
Overparameterized neural networks can fit arbitrary labels and random inputs \citep{zhang2017understanding}. However, networks also tend to learn simple, “shared” structure before fitting noisy samples \citep{arpit2017closer}, which has inspired methods that can leverage this tendency to train networks on very noisy data \citep{han2018coteaching}. Memorization can also be beneficial for generalization, and influence-based approaches suggest that memorization of atypical examples can improve performance on rare subpopulations \citep{feldman2020does,feldman2020what}.

\paragraph{Loss geometry, sharpness, and Fisher information.}
A significant body of work relates generalization to the local geometry of the loss landscape. Flatter minima have long been hypothesized to generalize better \citep{hochreiter1997flat}, and empirical studies have connected sharp minima to generalization gaps \citep{keskar2017large, jiang2020fantastic}.
Studies of the Hessian spectrum have shown that trained networks typically exhibit a small set of stiff directions along with a bulk of near-zero eigenvalues \citep{sagun2017eigenvalues,papyan2018full,ghorbani2019investigation}. The Fisher information provides a related local metric \citep{amari1998natural,martens2020new} which does not rely on a loss function to measure model sensitivity. Furthermore, there is evidence that flat directions connect minima more globally \citep{mode-con, pmlr-v80-draxler18a, frankle-LMC, carlo2026entropic}. Fisher information has been used to identify parameter directions important for previously learned tasks in continual learning \citep{kirkpatrick2017overcoming} and to study early phase training dynamics  \citep{jastrzebski2021catastrophic}.

\paragraph{In-context learning and task overfitting.}
Regression tasks have become a standard testbed for studying in-context learning \citep{garg2023transformerslearnincontextcase,akyurek2023what,vonoswald2023transformers}. More recent work shows that the emergence of in-context learning depends strongly on task diversity and training duration, such that transformers may first generalize to held-out tasks and later overfit by memorizing the finite set of training tasks \citep{raventos2023pretraining,singh2023the,nguyen2025differential}. Further work localizes the memorization to the MLP layers in the transformer \citep{singh2023the,nguyen2025differential}.

\paragraph{Benign overfitting and data coarse-graining.}
Our theoretical model studies a simplified setting of high-dimensional linear regression. In work on benign overfitting \citep{bartlett20,hastie2022surprises}, fitting pure noise can increase variance, and removing uninformative feature directions after fitting can reduce risk in finite samples. Our model isolates a ``barcode'' block that is uninformative across the population but is correlated with potential label noise. Data coarse-graining work shows that discarding data dimensions can improve predictive performance when the discarded degrees of freedom primarily encode noise \citep{nguyen26}.

\paragraph{Unlearning.} The de-memorization phenomena we study is distinct from unlearning research. Unlearning refers to the procedure of forcing a neural network to forget specific information on which it was trained, and is usually studied in the context of data privacy concerns or data which might produce harmful behavior in the model \citep{cao2015towards,ginart2019making,guo2020certified}. The type of model editing that we investigate does not support fine-grained deletion of information required for unlearning.

\section{Experiments}
To explore the interplay between memorization and loss geometry, we force models to memorize a subset of their training data using two variants of the standard CIFAR-10 classification task \cite{krizhevsky2009learning}. 

First, to study input noise, we augment the CIFAR-10 training set with 500 synthetic images (1\% of the training set). Following standard protocols for injected noise (see, e.g., Ref~\cite{zhang2017understanding}), each synthetic image is drawn from a Gaussian distribution whose mean and variance match the statistics of a randomly chosen CIFAR-10 class, but is assigned the label of an independently chosen random class. Because the input statistics and labels are independent, the network cannot rely on generalizable features, forcing it to explicitly memorize these inputs.

Second, to study label noise, we randomly corrupt 10\% of the training labels (see, e.g., Ref~\cite{garg2024memorization}). Each corrupted label is reassigned uniformly among the nine incorrect classes, forcing the network to explicitly memorize these modified samples.

In both settings, we train a ResNet-20 \citep{he2016deep} to 100\% training accuracy (details in Appendix~\ref{sec:training}). Following \citet{merullo2025memorization}, we compute and diagonalize each layer's Fisher information matrix and evaluate performance after projecting weights onto the top-$k$ subspaces. Unlike Ref~\cite{merullo2025memorization}, which relies on a K-FAC approximation, we compute the exact per-layer Fisher matrix. The Fisher information measures the sensitivity of the network's output to parameter perturbations,
\begin{equation}
\widehat{\mathbf{F}}(\boldsymbol{\theta}^{*}) = \frac{1}{N} \sum_{i=1}^{N} \sum_{c=1}^{C}\, p_{\boldsymbol{\theta}^{*}}(c\mid\mathbf{x}_i)\, \mathbf{s}_{ic}(\boldsymbol{\theta}^{*})\mathbf{s}_{ic}(\boldsymbol{\theta}^{*})^{\top}, \quad \mathbf{s}_{ic}(\boldsymbol{\theta}^{*}) \equiv\left.\nabla_{\boldsymbol{\theta}}\log p_{\boldsymbol{\theta}}(c\mid\mathbf{x}_i)\right|_{\boldsymbol{\theta}=\boldsymbol{\theta}^{*}}
\end{equation}
It is the Hessian of the Kullback-Leibler divergence, $D_\text{KL}(\theta||\theta')\approx\frac{1}{2}(\theta-\theta')^\top F(\theta)(\theta-\theta')$ for $\theta'\approx\theta$. We will show later that the dataset $D$ over which we compute the expectation can significantly influence the results of this procedure. 

\subsection{Which layers store memorized information?}\label{sec:gaussimage}
\begin{figure}
    \centering
    \includegraphics{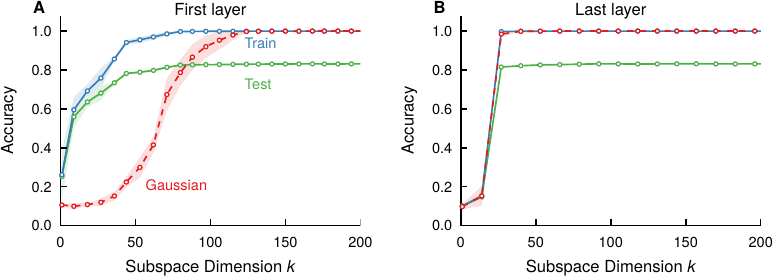}
    \caption{ \textbf{ Layer-dependent decoupling of memorization and generalization via Fisher projection: memorized Gaussian images.}  Classification accuracy as a function of the number of retained top Fisher eigenvectors $k$, computed over the full training set (CIFAR-10 and Gaussian images) for a single layer of a ResNet-20. Green: CIFAR-10 test; blue: CIFAR-10 train; red: memorized Gaussian samples. \textbf{A}) In the first layer, CIFAR-10 train and test accuracy rise in tandem and reach a maximum at $k\approx 45$, while accuracy on memorized Gaussian samples remains at $\approx\!30\%$, demonstrating selective de-memorization at negligible cost to generalization. \textbf{B}) In the last layer, all three curves rise sharply even for very small $k$, with no separation between them at any point, indicating that memorized and structured representations have merged and can no longer be separated.}
    \label{fig:bulk_fisher_gaussian}
\end{figure}

Figure \ref{fig:bulk_fisher_gaussian} illustrates the effect of this procedure on the first and last layers of a network trained on the Gaussian data. We compute the Fisher information over the entire training set, including both CIFAR and Gaussian images.

In the first layer, the model's accuracy on standard CIFAR-10 images (both train and test) rises quickly as a function of the eigenspace dimension $k$, outpacing its accuracy on the Gaussian data. At an eigenspace dimension of $k\approx45$, the model achieves near-maximum accuracy on the CIFAR-10 dataset while correctly classifying less than 30\% of the Gaussian samples. The accuracy on both the CIFAR-10 train and test sets rises in tandem, showing that this procedure separates the memorized data distribution from the learned distribution of CIFAR-10 images. 

Applying this projection to the final layer results in markedly different behavior. Accuracy across all three partitions (CIFAR-10 train, CIFAR-10 test and Gaussian) climbs simultaneously to full-model accuracy as the subspace dimension $k$ increases. By this layer, the model's representations of the two data distributions have converged, precluding selective removal. Taken together with the first layer, this indicates that the signature of memorization is layer-dependent, restricting de-memorization interventions to specific layers.

We also note that nearly all the relevant predictive information in the final layer's weights is concentrated within the span of the top \char`~5\% of its eigendirections.

\subsection{Memorization and generalization}\label{sec:labelflips}
In the Gaussian setting, de-memorization simply changes the model's predictions on unstructured inputs. When the model has memorized mislabeled examples, however, a correct answer exists, and we can ask whether de-memorization recovers it. To answer this question, we turn to the second setting, where we flip the labels of 10\% of the training set. When we ``de-memorize'' in this setting, we can ask how often the true label is recovered. 

\begin{figure}
    \centering
    \includegraphics{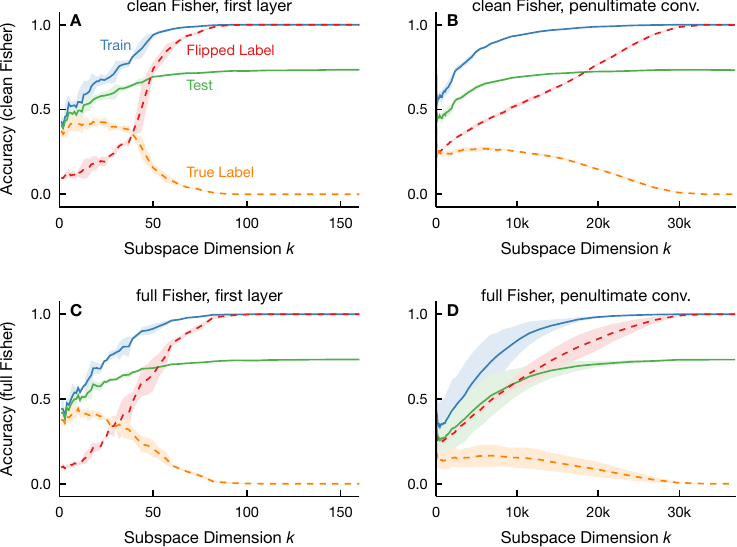}
\caption{ \textbf{ Layer-dependent decoupling of memorization and generalization via Fisher projection: noisy labels.} Classification accuracy as a function of retained top Fisher eigenvectors $k$ for a ResNet-20 trained on CIFAR-10 with 10\% label noise. Blue: train accuracy; Green: test accuracy; Red: accuracy on poisoned samples under their (wrong) training labels; Orange: accuracy on poisoned samples under their true labels. \textbf{A\&B)} Fisher computed on clean (non-corrupted) data only. \textbf{C\&D)} Fisher computed on the full training set. \textbf{First layer}, \textbf{A\&C)}: top Fisher eigenspaces treat memorized and clean samples differently under both Fisher variants. \textbf{Penultimate convolutional layer}, \textbf{B)}: with the clean Fisher, de-memorization is successful — the red curve has not reached its maximum by the time train and test accuracy do — and the model partially recovers the true labels of label-flipped samples, with the orange curve rising well above chance. \textbf{Penultimate convolutional layer}, \textbf{D)}: with the full Fisher, decoupling becomes weaker, highlighting the role of dataset choice in computing the Fisher.}
    \label{fig:labelnoise-layerwise}
\end{figure}

In Fig~\ref{fig:labelnoise-layerwise}, we show the results of the projection procedure for two representative layers (for all other layers, see Appendix Figs~\ref{fig:all-layers-clean} and \ref{fig:all-layers-full}). In the top row, we compute the Fisher information over a dataset of only clean examples and in the bottom row we compute the Fisher over a subset of the full training data. For the first layer (left column), we see similar behavior for both Fisher sets: the top subspace treats the data with real labels differently than the data with flipped labels. Although in this case the projection procedure does not completely remove the memorized data, we note that accuracy on the true labels of the flipped samples (orange curve) improves well above chance in the top Fisher subspace. 

In the right column, we consider the penultimate convolutional layer, the third-to-last layer in the network. In this layer, the data on which the Fisher is computed matters significantly: when we use only clean samples, the top eigenspace again treats the two types of data differently, but when we use samples from the full training set, including those with noisy labels, the accuracy curves move closer together. This highlights a fundamental property of the Fisher information: its geometric structure is highly dependent on the data distribution over which it is evaluated. 

\begin{figure}
    \centering
    \includegraphics{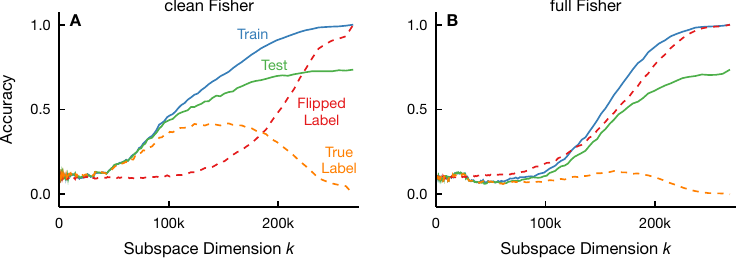}
    \caption{\textbf{Memorization and generalization across layers.} Results of using a block-diagonal approximation to estimate the Fisher information across all model parameters in the noisy-label setting. \textbf{A)} The Fisher is computed on training data with clean labels only. The top eigenspace selectively captures information about the true classification task. We also see non-monotonic behavior in the performance on the true labels of memorized samples (orange dashed line). \textbf{B)} The Fisher is computed on a sample from the full training dataset, including samples with corrupted labels. In this case, top eigenspaces mix information across tasks, emphasizing the data-dependence of the Fisher.}
    \label{fig:block_diag_labelnoise}
\end{figure}

What are the implications of the layer-wise analysis for the geometry of the full Fisher? Although computing the full Fisher is unwieldy for these models, we can use a block diagonal approximation to gain some insight into how layer-wise phenomena accumulate across the model. In Fig~\ref{fig:block_diag_labelnoise}, we show the results of the projection scheme for a block diagonal Fisher computed on clean data only (A) and a sample of the full training set (B).

When computing the Fisher matrix exclusively on data with clean labels (Fig~\ref{fig:block_diag_labelnoise}A), the top eigenspaces selectively capture information about the true classification task. In these dominant subspaces, the network starts to ignore the corrupted labels it was forced to memorize during training. However, when we compute the Fisher on a sample from the full training dataset (Fig~\ref{fig:block_diag_labelnoise}B), this geometric separation breaks down. Without the clean data filtering, the top eigenspaces mix information across the generalization and memorization tasks---in order to see any separability, we need to turn to the individual layers (Fig~\ref{fig:labelnoise-layerwise}).

\subsection{In-context learning \label{sec:icl}}

We next investigate memorization in an in-context learning (ICL) setting \citep{NEURIPS2020_1457c0d6}. A growing body of work has developed in-context learning of linear regression as a simple model of ICL amenable to quantitative study \citep{garg2023transformerslearnincontextcase, raventos2023pretraining, goddard2025when, hill2025transformersdontincontextlearn}. One important feature of ICL captured by the linear regression model is that ICL behavior can be \textit{transient}, emerging early on during training before the model memorizes the set of training tasks \citep{singh2023the, raventos2023pretraining,nguyen2025differential}. Here, we investigate how the model encodes these memorized tasks, and show that we can recover at least some of the transient ICL performance using Fisher projection. 

\begin{figure}
    \centering
    \includegraphics{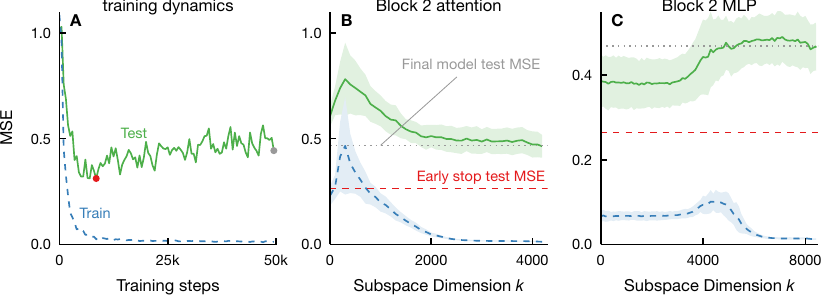}
    \caption{\textbf{From memorization to generalization, in-context.} \textbf{A)} Transient ICL training dynamics. The three-block transformer first improves performance on held-out tasks, before overfitting to training tasks. \textbf{B)} Fisher projections for the attention layer, second transformer block. All projections decrease performance. Shaded region shows standard deviation across 16 models. \textbf{C)} Fisher projections for the MLP layer, second transformer block. Top Fisher subspaces improve task generalization (green) at the cost of performance on memorized tasks (blue).}
    \label{fig:icl}
\end{figure}

We train a three-block GPT-2 style transformer on contexts $C_\tau = (x_1,y_1, x_2,y_2,\dots,x_n)$ with $y_i = w_\tau^Tx_i$. The model must learn to effectively infer the latent regression ``task'' vector $w_\tau$ to predict $y_n$. Following Ref~\cite{raventos2023pretraining}, training contexts are sampled uniformly from a set of $M$ task vectors, $\{w_\tau\}_{\tau=1}^M$. As shown in Fig~\ref{fig:icl}A, when $M$ is small, performance on held-out tasks first improves until the model overfits to the training tasks. We emphasize that the model never directly sees the task vectors, and so this setting, we investigate memorization of \emph{latent} ``concepts'' rather than the direct memorization of the input data.

We next compute the Fisher information for each component in the model, and project the model parameters onto top-$k$ subspaces. In Fig~\ref{fig:icl}B and C, we show the results for the second transformer block's attention and MLP layers (See Appendix Fig~\ref{fig:icl-all} for all layers). We see that in the attention layer, projecting onto top subspace only hurts model performance, on both the memorized training tasks and held-out test tasks. However, the same projection procedure applied to the MLP layer yields a substantial improvement in test performance at the cost of performance on the memorized tasks. We note that prior work also identified the MLP as the component of a transformer responsible for memorization \citep{singh2023the,nguyen2025differential}; our analysis further suggests that this memorized information is stored in low-Fisher-eigenvalue parameter components. Moreover, removing memorized information can partially restore in-context learning based task generalization, closing roughly half the gap to the best early-stopped performance.

\section{Contributions of Individual Samples}

\subsection{Sample Contributions to the Eigenspectrum}

To better understand how individual data samples contribute to generalization and the structure of the Fisher information, we examine the participation ratios associated with individual samples in the eigenspectrum of the (layerwise) Fisher. To get some intuition, we can express the Fisher information as an average over per-sample Fishers: $\frac{1}{n} \sum_{i=1}^n F_i$. Then, diagonalizing the Fisher as $F = \sum_{k=1}^p \lambda_k \vec{u}_k  \vec{u}^\intercal_k$, we define the contribution of sample $i$ to mode $k$ as $c_i^k = \vec{u}^\intercal_k F_i \vec{u}_k$. The participation ratio for mode $k$ is then $\Psi^k = (\sum_i{c_i^k})^2/(n\sum_i{(c_i^k)^2})$. This ratio is a measure of the fraction of data samples that contribute to mode $k$. For example, when $c_i^k=1$ for all $i$, $\Psi^k=1$ and all samples contribute (equally). On the other hand, when only a single sample contributes,  $\Psi^k=\frac{1}{n}$ and the participation ratio is suppressed by the size of the dataset. The restricted participation ratio $\Psi^k_{\textbf{CIFAR}} \equiv  (\sum_{j \in \textbf{CIFAR}}{c_j^k})^2/(n\sum_i{(c_i^k)^2})$ restricts the numerator to only the contributions of the CIFAR samples, and this convention is used equivalently for  $\Psi^k_{\textbf{Gauss}}$,$\Psi^k_{\textbf{corrupted}}$, and $\Psi^k_{\textbf{uncorrupted}}$. Note that $\Psi^k \neq \Psi^k_\textbf{CIFAR} + \Psi^k_\textbf{Gauss}$ generically because of cross-terms.

\begin{figure}
    \centering    \includegraphics[width=\linewidth]{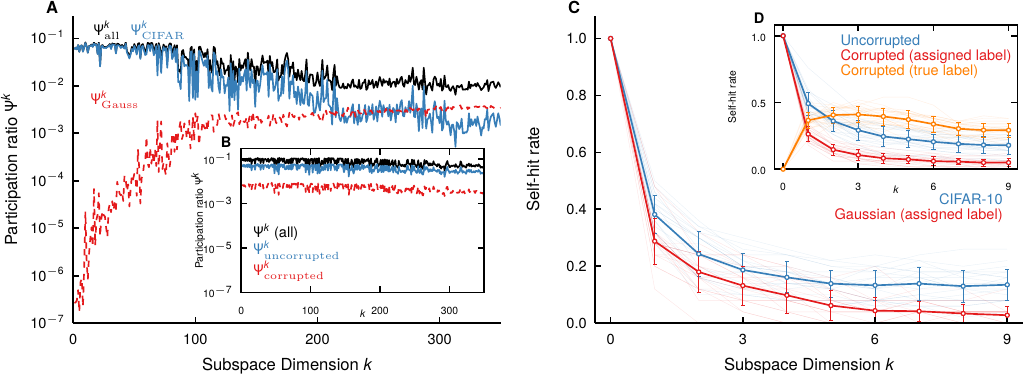}
    \caption{\textbf{Single-sample contributions and projections.} Participation ratios for \textbf{A)} input-noise case and \textbf{B)} label-noise case. \textbf{A)} suggests that structured CIFAR images primarily contribute to the top of the eigenspectrum, while noisy samples primarily contribute to the bottom of the eigenspectrum. This separation does not occur in \textbf{B)}. \textbf{C)} We plot the self-hit rate of correct sample classification after projecting out the top-k subspace of the corresponding single-sample Fisher information (`sample projection') in the input-noise case, and \textbf{D)} shows the same statistic for the label-noise case. Noisy samples are less likely to be recovered than their counterparts.}
    \label{fig:single_sample}
\end{figure}

Figure~\ref{fig:single_sample}A and B display first-layer participation ratios across the eigenspectrum. In the input-noise experiments (Fig~\ref{fig:single_sample}A), we see total participation ratios drop by roughly one order of magnitude from top to bottom modes. Importantly, the restricted ratios show that clean CIFAR-10 images dominate the top modes, whereas memorized Gaussian noise is relegated to the bottom. These findings suggest that the top eigenmodes capture information about structured, generalizing features in the data, while the bottom portion corresponds to memorized, idiosyncratic information. By contrast, in the noisy-label experiments (Fig~\ref{fig:single_sample}B), we see that corrupted and uncorrupted samples contribute in a constant ratio across the entire spectrum.

\begin{table*}[htbp]
\centering
\caption{Bulk accuracy after sample projection. The $k=0$ column is the matched unchanged-weight BatchNorm control (see Appendix \ref{app:single_sample} for results on other Fisher sources).}
\label{tab:trunc_table}
\begingroup
\vspace{1mm}
\scriptsize
\setlength{\tabcolsep}{2pt} 
\renewcommand{\arraystretch}{1.12} 

\begin{tabular}{@{}lccccc@{}c@{}lccccc@{}}
\multicolumn{6}{c}{\textbf{Gaussian model}}
& \hspace{1.5em} &
\multicolumn{6}{c}{\textbf{Noisy-label model (10\% noise)}} \\[2pt]

\cline{1-6}\cline{8-13}
& $k=0$ & $1$ & $2$ & $4$ & $9$
&&
& $k=0$ & $1$ & $2$ & $4$ & $9$ \\
\cline{1-6}\cline{8-13}

\multicolumn{6}{l}{\textit{Fisher source CIFAR-10 (\%)}}
&&
\multicolumn{6}{l}{\textit{Fisher source uncorrupted (\%)}} \\
\cline{1-6}\cline{8-13}

CIFAR-10 train (5k samples)
& 100.0 & 99.9 & 99.9 & 99.9 & 99.8
&&
CIFAR-10 train (5k samples)
& 100.0 & 100.0 & 100.0 & 99.9 & 99.9 \\

CIFAR-10 test
& 84.3 & 83.9 & 83.7 & 83.4 & 83.1
&&
CIFAR-10 test
& 73.4 & 73.1 & 73.0 & 72.7 & 72.3 \\
\cline{1-6}\cline{8-13}

\multicolumn{6}{l}{\textit{Fisher source Gaussian (\%)}}
&&
\multicolumn{6}{l}{\textit{Fisher source corrupted (\%)}} \\
\cline{1-6}\cline{8-13}

Gaussian set
& 100.0 & 98.4 & 96.1 & 90.5 & 75.6
&&
Corrupted CIFAR (assigned labels)
& 100.0 & 100.0 & 100.0 & 99.9 & 99.8 \\
\cline{1-6}\cline{8-13}
\end{tabular}

\endgroup
\end{table*}

\subsection{Single-Sample Projections}
We also investigate single-sample deletion by projecting out top eigenvectors of single-sample Fisher information matrices, as we show in Fig~\ref{fig:single_sample}C and D. We refer to the process of calculating a sample's single-sample Fisher information and projecting out top eigenvectors of this Fisher as `sample projection.' 

As depicted in Fig~\ref{fig:single_sample}, classification accuracy after projection is higher for clean CIFAR-10 images than for Gaussian inputs, and this gap widens as more eigenvectors are projected out. Likewise, in the label-noise experiments, not only are uncorrupted samples more resilient to projection, corrupted samples also recover their true labels more often than their assigned false labels. True-class recovery is non-monotonic in the number of removed subspace dimensions. Finally, we note that sample projections minimally degrade samples besides the one targeted: the exception to this trend is significant degradation of performance on the Gaussian set after projecting out a Gaussian sample (see Table \ref{tab:trunc_table}).

\section{A barcode model of label-noise memorization}\label{sec:barcode}

Our experiments demonstrate that projecting model weights onto top Fisher modes degrades the model's recall of memorized samples while preserving generalization to clean data. Curvature-based editing~\cite{merullo2025memorization} and coarse-graining of uninformative features~\cite{bartlett20,nguyen26} offer partial accounts of this effect, but closed-form analyses for discrete label flips remain unavailable. To illustrate this mechanism, we develop an idealized linear model that yields exact expressions for the risk reduction from projection and reveals a regime in which an interior-optimal projection rank emerges.

\begin{figure}
  \centering
  \includegraphics[width=\linewidth]{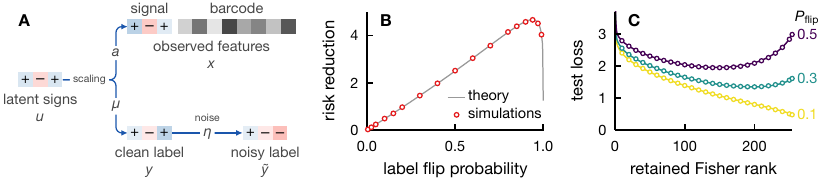}
  \caption{\textbf{Finite-rank linear model illustrates the risk gain from Fisher projection.}
  \textbf{A)}~Generative model. A latent sign vector $u$ is drawn uniformly from $\{-1,+1\}^r$; anisotropic scaling by amplitude vectors $\mu$ and $a$ produces signal features $s$ and clean label $y$, respectively. The observed features $x$ result from concatenating the signal $s$ and the uninformative barcode features $z$, modeled as an isotropic Gaussian vector in $d$ dimensions. The training label $\tilde y$ is the clean label $y$ with random per-dimension sign flips, with a random sign vector $\eta$.
  \textbf{B)}~Risk gain from removing the barcode block post hoc \emph{vs} label flip probability $P_{\mathrm{flip}}$, demonstrating quantitative agreement between theory (solid curve, Eq~\eqref{eq:DeltaR}) and simulations (circles; error bars are smaller than the circle size).
  \textbf{C)}~Prediction MSE \emph{vs} retained Fisher rank for three flip rates (see legend). Two of four signal dimensions have population variance below the spectral separation threshold and therefore mix with the barcode bulk. In both theoretical prediction  (solid curve, Eq~\eqref{eq:rmixed}) and simulations (circles), an interior optimum emerges as the flip rate increases. 
  Shown results are for $(n,r,d)=(500,4,250)$ with $200$ replicates.}
  \label{fig:barcode}
\end{figure}

Fig~\ref{fig:barcode}A depicts the generative model. For each input-label pair $(x,y)$, we generate a sign vector $u\sim \mathrm{Uniform}(\{-1,+1\}^r)$. The feature vector $x$ consists of informative features $s$, obtained from anisotropically scaling the sign vector, and barcode features $z$, modeled as an isotropic Gaussian vector; that is,
\begin{equation}\label{eq:feature}
x = \genfrac{[}{]}{0pt}{1}{s}{z}\in\R^{r+d}
\quad\text{with}\quad s_i = \mu_i u_i 
\quad\text{and}\quad z\sim \mathcal{N}(0,I_d),
\end{equation}
where $\mu$ is a fixed nonnegative vector. 
The clean label $y$ is a scaled sign vector, and the corrupted training label $\tilde y$ is subject to random per-dimension sign flips, 
\begin{equation}\label{eq:clean-target}
y_i = a_i u_i \quad\text{and}\quad
\tilde y_i = \eta_i y_i
\end{equation}
where $a$ denotes a fixed nonnegative vector, $\eta\in\{-1,+1\}^r$ a random vector with $P(\eta_i\!=\!-1)=\rho_i$. The probability that at least one coordinate flips is $P_{\mathrm{flip}}=1-\prod_i(1-\rho_i)$. The corrupted label decomposes into a shrunk clean label and a noise term,
\begin{equation}\label{eq:corrupted-coordinate}
\tilde y_i=\bar\eta_iy_i+\xi_i,
\end{equation}
where $\bar\eta_i\equiv\EE[\eta_i]=1-2\rho_i$. The noise term is given by
\begin{equation}\label{eq:label-noise}
\xi_i = (\eta_i - \bar\eta_i)y_i
\quad\text{with}\quad
\EE[\xi_j\mid y]=0
\quad\text{and}\quad
\EE[\xi_i\xi_j\mid y]= \delta_{ij} a_i^2(1-\bar\eta_i^2).
\end{equation}

Stacking $n$ independent training samples yields the design matrix $X=\genfrac{[}{]}{0pt}{2}{S}{Z}\in\R^{(r+d)\times n}$, with informative-feature block $S\in\R^{r\times n}$, barcode block $Z\in\R^{d\times n}$, corrupted-label matrix $\tilde Y\in\R^{r\times n}$ and noise matrix $\Xi\in\R^{r\times n}$. 

We fit a linear model, $\hat y=\hat\Theta_s s+\hat\Theta_z z$, on the training data $(X,\tilde Y)$ and evaluate its prediction risk on a fresh sample $(x,y)$ from the \emph{clean} generative model, i.e., no corrupted label at test time.

Minimizing $\|\tilde Y-\Theta_s S-\Theta_z Z\|_F^2$ over coefficient blocks $\Theta_s\in\R^{r\times r}$ and $\Theta_z\in\R^{r\times d}$ gives
\begin{equation}\label{eq:thetaz}
\hat\Theta_z=\Xi\,\Pi_S^\perp Z^\top\left(Z\Pi_S^\perp Z^\top\right)^{-1},
\end{equation}
where $\Pi_S^\perp=I_n-S^\top(SS^\top)^{-1}S$ projects onto the complement of the informative-feature row space. This coefficient captures finite-sample correlation between the barcode design $Z$ and the centered label noise $\Xi$. It vanishes in the infinite-data limit since $z$ and $\xi$ are uncorrelated. 

We now calculate the effect of a post-hoc deletion of the barcode block. To do this, we consider the model $\hat y=\hat\Theta_s s$ instead of $\hat y=\hat\Theta_s s+\hat\Theta_z z$.\!\footnote{%
This procedure is not the same as deleting the barcode before fitting. Using both $S$ and $Z$ allows predictive and barcode features to compete, as both are correlated with the corrupted training labels in the finite-sample regime.} 
This intervention reduces prediction risk by
\begin{equation}\label{eq:DeltaR}
\Delta R
=\EE\left\|y-\hat\Theta_s s-\hat\Theta_z z\right\|^2
-\EE\left\|y-\hat\Theta_s s\right\|^2
=\tr\left(\Sigma_\xi\right)\frac{d}{n-r-d-1},
\quad\text{for}\quad n-r>d+1,
\end{equation}
where the expectation is over both the training set $(X,\tilde Y)$, on which the estimators depend, and a fresh test point $(x,y)$. 
This expression is equivalent to the variance that a $d$-dimensional linear regression with Gaussian covariates accumulates when fitting pure noise with $n-r$ effective samples.
The condition $n-r>d+1$ requires that the barcode dimension of the local model sit below the effective sample count; it does not constrain the total parameter count of a larger model. In Fig~\ref{fig:barcode}B, we depict the reduction in prediction risk, resulting from barcode deletion, over the full range of label flip probability $P_{\mathrm{flip}}$. We observe good agreement between the theoretical prediction, Eq~\eqref{eq:DeltaR}, and numerical simulations.

Crucially, equation~\eqref{eq:DeltaR} assumes a clean separation between the informative and barcode directions, but Fisher projection on a real network operates in the empirical eigenmodes of the local feature covariance, where the two can mix. When an informative feature has population variance $\smash{\mu_j^2}$ below the spectral separation threshold $1+\smash{\sqrt{d/(n-r)}}$, its empirical eigenvector merges into the barcode bulk~\cite{baik05}. In this case, discarding low-eigenvalue directions sacrifices predictive information as well as memorized noise. Whether a finite-rank intervention benefits then depends on the regime. At low flip rates, the prediction-risk curve is monotone in retained rank and full rank is best. At higher flip rates, on the other hand, an interior-optimal rank emerges, set by the trade-off between memorization variance and informative-feature leakage. Fig~\ref{fig:barcode}C depicts this transition in the linear model; Appendix~\ref{sec:fisher} develops the partial-overlap theory underlying the solid curves.

\section{Discussion and Outlook}\label{sec:discuss}

Across a range of settings, our work shows that the memorized and generalized information within a neural network is separated by its Fisher geometry. In particular, retaining only the top of the eigenspectrum of the Fisher in certain layers induces de-memorization. Our findings suggest a picture of memorization which is both diffuse and localizable: single-layer interventions allow us to de-memorize noisy samples without degrading task performance on structured data, but we also observe an accumulation of layerwise phenomena in the Fisher information of the entire model.

Prior work has shown that adversarial competition exists between memorizing and generalizing circuits when language models are trained to perform in-context learning \cite{nguyen2025differential,gibson2026distinctmechanismsunderlyingincontext}. Our work suggests that, in some cases, de-memorization can improve generalization without the need for retraining. Interference between the model's memorizing and generalizing components seem to ``hide'' its full generalization capacity: by projecting out a certain fraction of components, we improve performance on tasks which benefit from generalization.  

At the scale of the entire network, we find that the Fisher-geometric signature of memorization exhibits a strong data-dependence: it is easier to disentangle memorization from generalization when the Fisher is computed on a denoised version of the training data. The collapse of this disentanglability occurs in particular layers of the network, rather than within each layer.

We remark on two surprising features of the above phenomena from the perspective of information geometry. First, we have related memorization to a linear decomposition of parameter space. Second, we identify memorizing/generalizing subspaces by means of a local geometric quantity (the Fisher), but the projection procedure we perform is a non-local transformation: zeroing out eigenvectors can cause large perturbations to the network's parameters. Prior work on loss-landscape geometry suggests that, in high-dimensional optimization problems, many soft directions exist in which large parameter-space perturbations correspond to small function-space perturbations, forming a so-called hyperribbon~\cite{Quinn_2023}. One possibility is that the nonlocal projection operation in parameter space corresponds to a small perturbation in the space of models. We leave an investigation of this possibility to future work.

Our results should be interpreted as an exploration of the geometric structure of memorization and generalization, not as a scale-independent prescription for model editing. Our empirical analysis uses ResNet-20 classifiers on CIFAR-10 variants and small transformers trained on synthetic in-context regression. The layerwise separation we observe may change qualitatively in larger models. Our vision experiments focus on simple object classification with controlled inputs and label noise; other data modalities may induce different forms of memorization---particularly when the memorized content is semantic, compositional or distributed across examples. We also emphasize that our analysis ignores cross-layer correlations, which may contribute meaningfully to memorization. 

Our work contributes to a growing body of research which supports the possibility of ``intelligent yet ignorant'' models: models which can reason, but which have no knowledge of particular information. In particular, we have provided a number of examples where learning a task and memorizing data can be decoupled, and we demonstrate a geometric mechanism for this decoupling. 

\begin{ack}
VN was supported in part by grants from the NSF (DMS-2235451) and Simons Foundation (MPS-NITMB-00005320) to the NSF-Simons National Institute for Theory and Mathematics in Biology (NITMB). DJS was partially supported by a Simons Fellowship in the MMLS, a Sloan Fellowship, and the National Science Foundation, through the Center for the Physics of Biological Function (PHY-1734030). VN and DJS also acknowledge funding from Coefficient Giving.
\end{ack}

\bibliography{references}

\appendix

\section{Additional Experimental Details}
\subsection{Model Training \label{sec:training}}
We train all Resnet-20 models for 200 epochs with an initial learning rate of 0.1, and quench the learning rate by a factor 10 at epoch 100 and epoch 150. We use a batch size of 128 and train the models using SGD with weight decay $10^{-4}$ and momentum 0.9. Each model is trained on a single GPU. The model training requires 1-2 GPU hours, and the Fisher projection experiments require $\sim$12 GPU hours per model on an 80GB GPU. We are careful to re-estimate batchnorm parameters after each projection.

The transformer used in section \ref{sec:icl} is a standard GPT-2 style transformer with learned positional embeddings, hidden dimension 32, and consists of three transformer blocks and a linear embedding from the regression dimension to the hidden dimension. We use 8 attention heads in each block, and dropout 0.1. We generate sequences of 20 $(x,y)$ pairs at training and test time, and $x\in\mathbb{R}^5$.

\section{Supplemental figures}

\begin{figure}[ht!]
    \centering
\includegraphics[width=\linewidth]{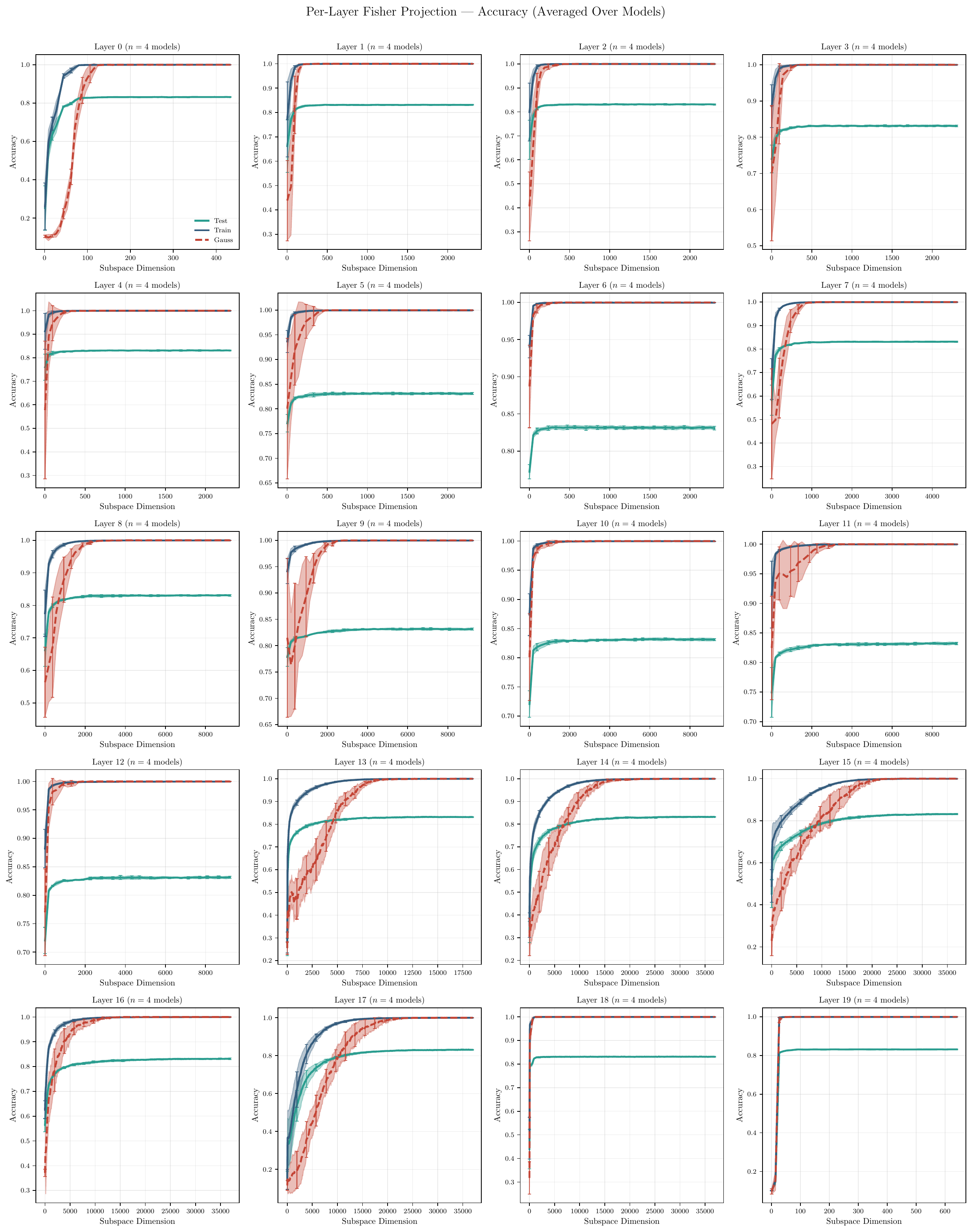}
    \caption{Layerwise Fisher projection experiments for the input-noise case.}
    \label{fig:appendix_gaussfisher}
\end{figure}

\begin{figure}[ht!]
    \centering
    \includegraphics[width=1.0\linewidth]{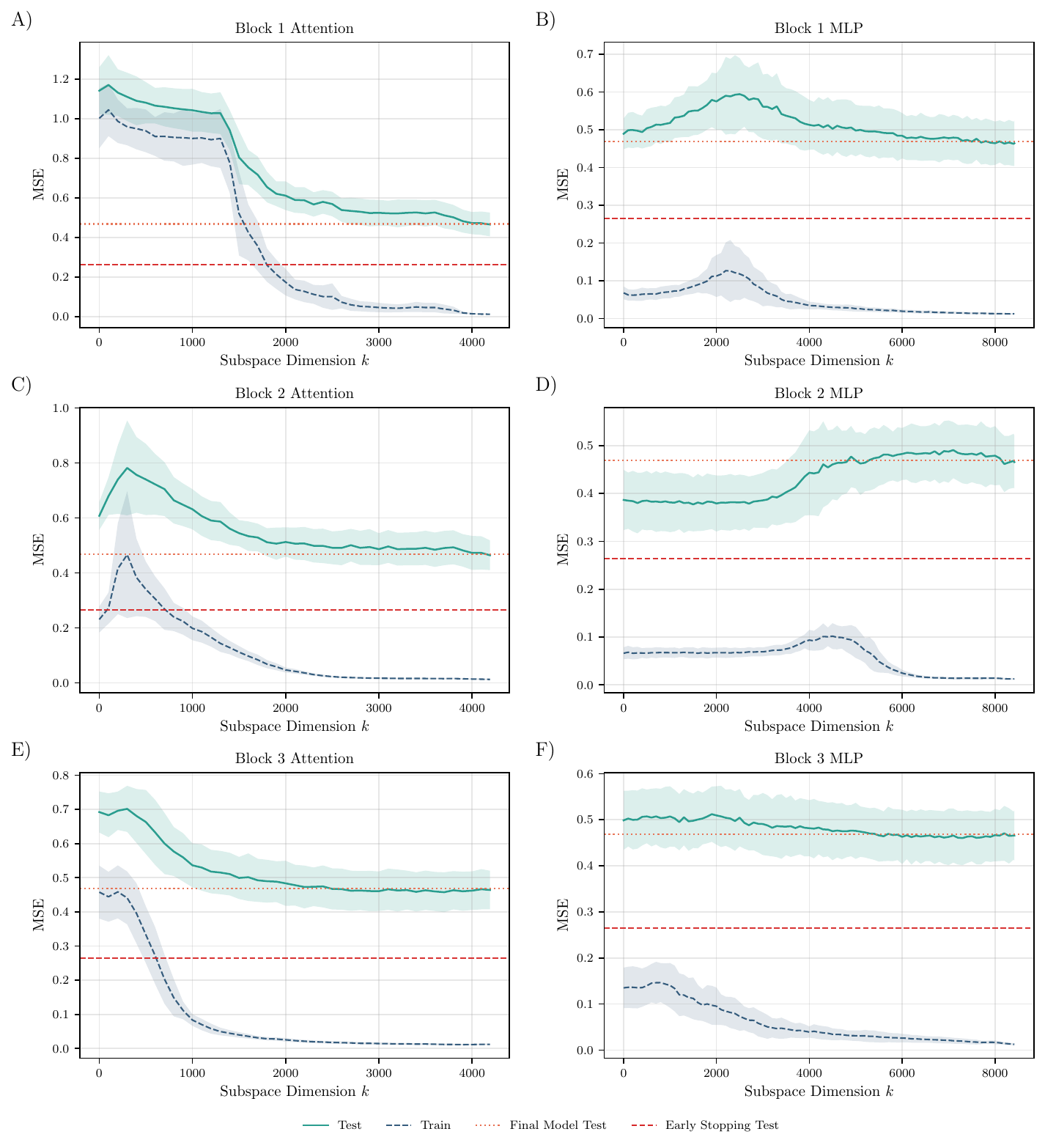}
    \caption{\textbf{Memorization concentrates in a single MLP.} Here, we show all transformer blocks for the model considered in Fig~\ref{fig:icl}. All layers except the Block 2 MLP perform worse on both train/test tasks when restricted to a top Fisher subspace, but in the Block 2 MLP test performance improves at the cost of performance on the training tasks.}
    \label{fig:icl-all}
\end{figure}

\begin{figure}[ht!]
    \centering
    \includegraphics[width=1.0\linewidth]{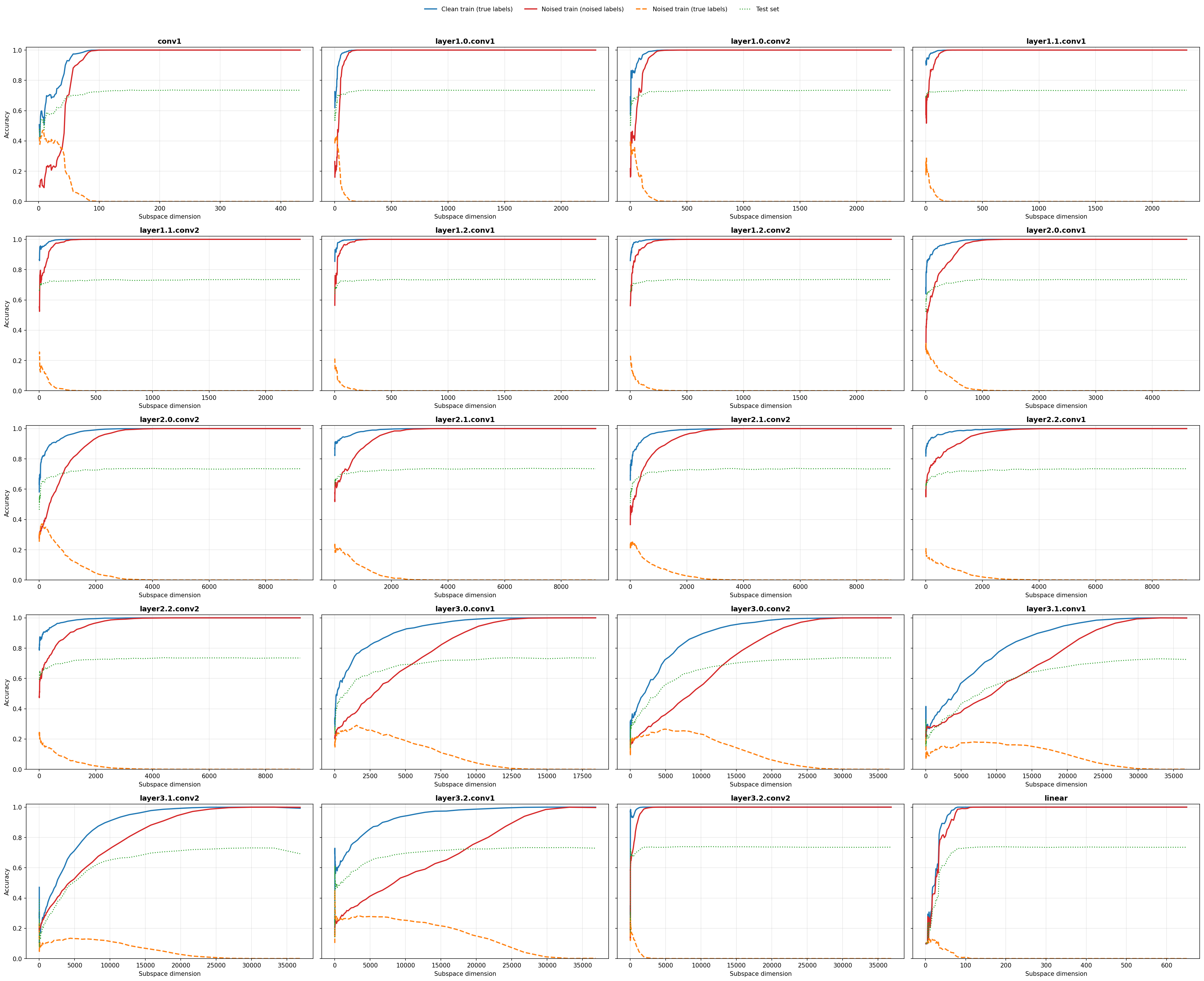}
    \caption{Fisher projections for every layer, in the noisy sample setting. Fisher computed on clean samples.}
    \label{fig:all-layers-clean}
\end{figure}

\begin{figure}[ht!]
    \centering
    \includegraphics[width=1.0\linewidth]{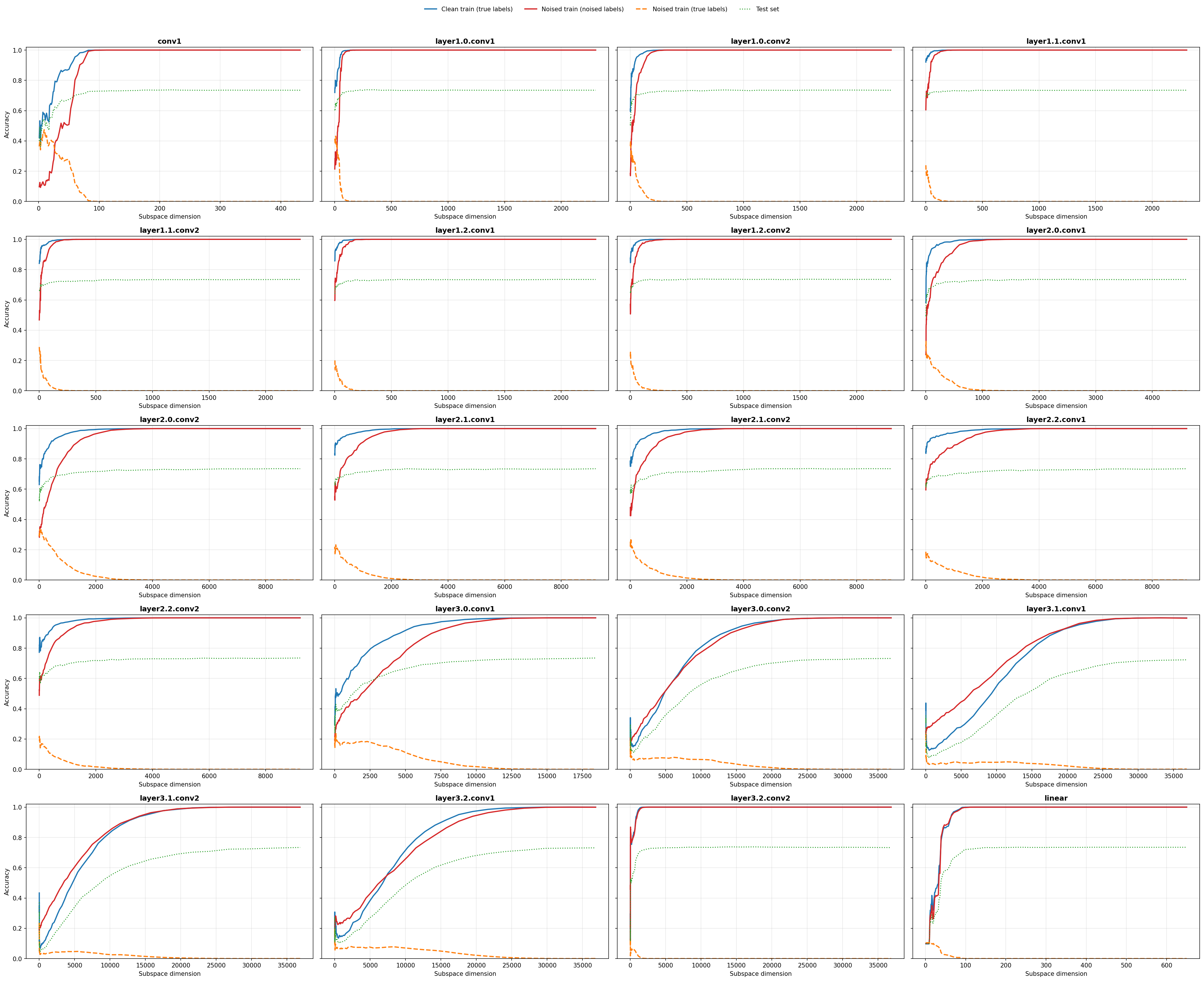}
    \caption{Fisher projections for every layer, in the noisy sample setting. Fisher computed on a subset of the full training set.}
    \label{fig:all-layers-full}
\end{figure}

\begin{figure}
    \centering
   \includegraphics[width=1.0\linewidth]{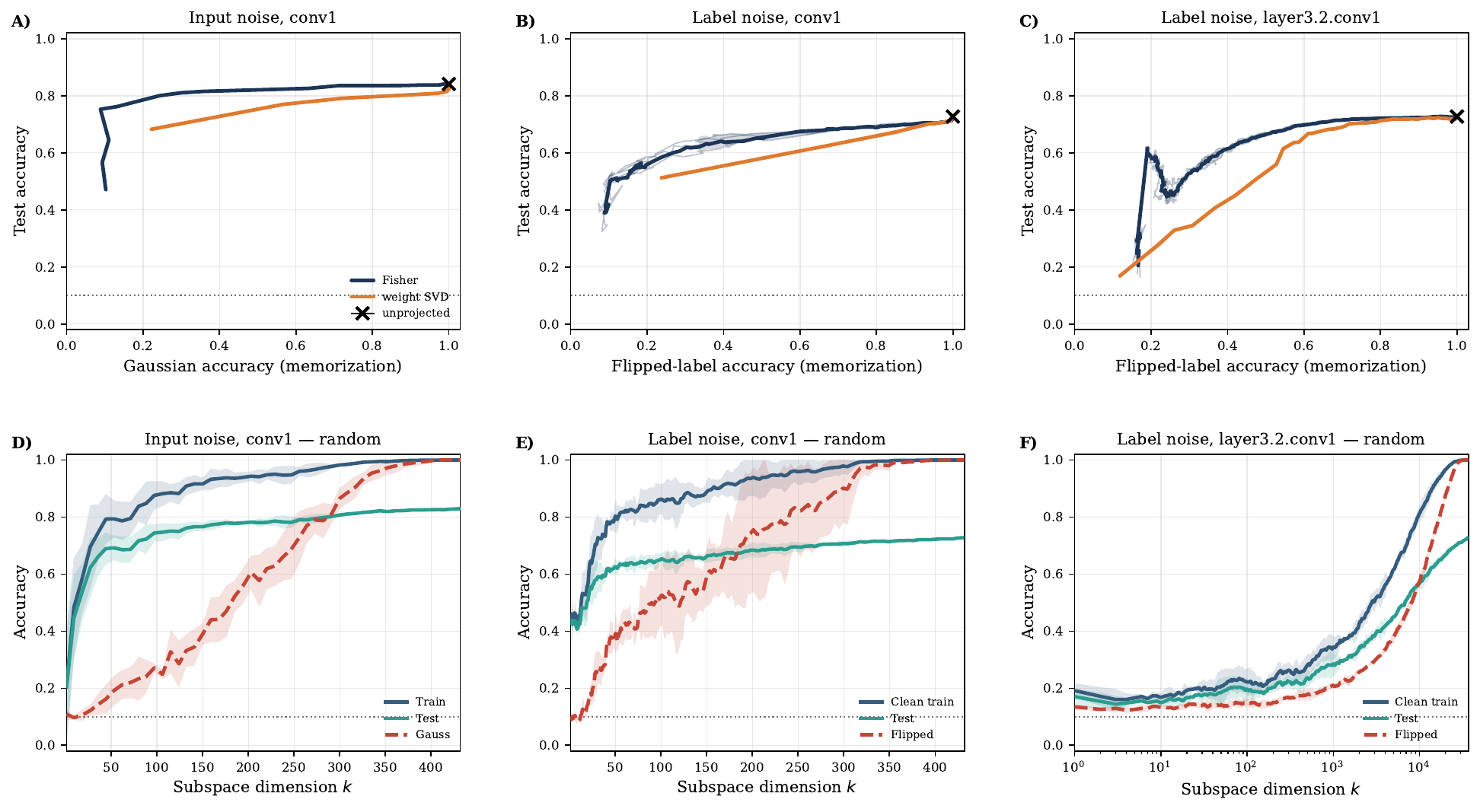}
    \caption{Control experiments for \textbf{A)-C)} SVD compression and \textbf{D)-F)} random basis projections.} 
    \label{fig:control_appendix}
\end{figure}

In Fig~\ref{fig:appendix_gaussfisher}  we show projections for all layers in the input-noise setting. In Fig~\ref{fig:icl-all}, we show all transformer blocks for our ICL setting. In Fig~\ref{fig:all-layers-clean} and Fig \ref{fig:all-layers-full}, we show projections for all layers in the label-noise setting, both for a clean Fisher and the Fisher of the full training set. In Fig~\ref{fig:control_appendix} we show the results of control experiments in which we project models iteratively onto an alternative basis to the Fisher eigenbasis. 

\eject
\clearpage

\section{Barcode model derivations}\label{app:barcode}

Here, we derive the results stated in \S\ref {sec:barcode}. We use the same notation as in the main text, $X = \genfrac{[}{]}{0pt}{1}{S}{Z}$,
$\tilde Y = D_{\bar\eta}Y + \Xi$ and
$\Pi_S^\perp = I_n - S^\top(SS^\top)^{-1}S$.
We write $\Sigma_\xi = \diag(a_1^2(1-\bar\eta_1^2),\ldots,a_r^2(1-\bar\eta_r^2))$
for the noise covariance defined in Eq~\eqref{eq:label-noise}.

\subsection{Estimating block coefficients}\label{app:ols}

We obtain the estimators for the block coefficients by minimizing the squared loss
\begin{equation}
(\hat\Theta_s,\hat\Theta_z) = \min_{(\Theta_s,\Theta_z)}L(\Theta_s,\Theta_z)
\quad\text{where}\quad
L(\Theta_s,\Theta_z)=\|\tilde Y - \Theta_s S - \Theta_z Z\|^2_F.
\end{equation}
At fixed $\Theta_z$, the optimal signal block reads
\begin{equation}
\hat\Theta_s(\Theta_z)=(\tilde Y - \Theta_z Z)S^\top(SS^\top)^{-1}.
\end{equation}
Substituting this expression in the loss yields
\begin{equation}
L(\hat\Theta_s(\Theta_z),\Theta_z) = \|(\tilde Y - \Theta_z Z)\Pi_S^\perp\|_F^2.
\end{equation}
Equating its derivative to zero gives
\begin{equation}
(\tilde Y - \hat\Theta_z Z)\Pi_S^\perp(Z\Pi_S^\perp)^\top = 0.
\end{equation}
When $Z\Pi_S^\perp$ has full row rank $d$ (i.e., $n\!-\!r \ge d$), this equation admits a unique solution
\begin{equation}\label{eq:thetaz-app}
\hat\Theta_z 
= \tilde Y\Pi_S^\perp Z^\top(Z\Pi_S^\perp Z^\top)^{-1}
= \Xi\Pi_S^\perp Z^\top(Z\Pi_S^\perp Z^\top)^{-1}, 
\end{equation}
The last equality results from the fact that $Y = D_a U= D_a D_\mu^{-1}S$ and $S\Pi_S^\perp=0$. Hence, we have derived Eq~\eqref{eq:thetaz}.

\subsection{Risk reduction from barcode deletion}\label{app:risk}

We now derive the risk gain in Eq~\eqref{eq:DeltaR},
\begin{align}
\Delta R
&= \EE\|y - \hat\Theta_s s - \hat\Theta_z z\|^2 - \EE\|y - \hat\Theta_s s\|^2\\
&= \EE[-2(y-\hat\Theta_s s)^\top\hat\Theta_z z] + \EE[\|\hat\Theta_z z\|^2] \\
&= \EE[\|\hat\Theta_z\|^2_F],\label{eq:gap-to-frob}
\end{align}
where the expectation is over the training set $(X,\tilde Y)$ and a fresh test point $(x,y)$. In the above, we used the fact that the test barcode $z$ is independent of all other random variables with $\EE[z]=0$ and $\EE[zz^\top]=I_d$.

We evaluate this expression in two steps. First, we condition on the design $(S,Z)$ and average over $\Xi$. Substituting Eq~\eqref{eq:thetaz-app} in Eq~\eqref{eq:gap-to-frob} yields 
\begin{align}
\Delta R &= \tr\EE\left[
\Xi\Pi_S^\perp Z^\top(Z\Pi_S^\perp Z^\top)^{-1}(Z\Pi_S^\perp Z^\top)^{-1}
Z \Pi_S^\perp \Xi^\top
\right]
\\
&= \tr
\EE\left[
\EE[\Xi^\top\Xi\mid S,Z]
\Pi_S^\perp Z^\top(Z\Pi_S^\perp Z^\top)^{-1}(Z\Pi_S^\perp Z^\top)^{-1}
Z \Pi_S^\perp 
\right]
\\
&= \tr(\Sigma_\xi)\tr
\EE\left[(Z\Pi_S^\perp Z^\top)^{-1}\right].
\end{align}
In deriving the above, we use $\EE[\Xi^\top\Xi\mid S,Z]=\tr(\Sigma_\xi)I_n$.

Second, we average over the Gaussian barcode $Z$. The projection $\Pi^\perp_S$ zeros out the $r$ signal rows, leaving $m=n-r$ degrees of freedom. Since $Z$ has iid zero-mean, unit-variance Gaussian entries, independent of $S$, the matrix $Z\Pi_S^\perp Z^\top$ follows a $\mathrm{Wishart}_d(I_d,m)$ distribution.\!\footnote{%
To see this, let $U\in\R^{n\times m}$ have orthonormal columns spanning the range of $\Pi_S^\perp$, so $\Pi_S^\perp = UU^\top$. Then $Z\Pi_S^\perp Z^\top = (ZU)(ZU)^\top$ and $ZU\in\R^{d\times m}$ has iid standard Gaussian entries.%
}
For $m > d+1$, the inverse-Wishart mean gives $\EE[(Z\Pi_S^\perp Z^\top)^{-1}] = I_d/(m-d-1)$ (see, e.g., Ref~\cite{anderson03}). It immediately follows that
\begin{equation}
\Delta R = \tr(\Sigma_\xi)\,\frac{d}{n-r-d-1},
\end{equation}
which concludes our derivation of Eq~\eqref{eq:DeltaR}.

\subsection{Partial-overlap prediction MSE}\label{sec:fisher}

The risk gain in Eq~\eqref{eq:DeltaR} assumes clean separation between signal and barcode directions. The residual barcode covariance has upper sample-bulk edge $(1+\sqrt{d/(n-r)})^2$, while a signal coordinate separates from that bulk only when its population variance satisfies $\mu_j^2>1+\sqrt{d/(n-r)}$. Below this population threshold, empirical eigenvectors mix signal and barcode directions, and the projection rank $K$ trades off signal loss against noise removal~\cite{baik05}.

Let $XX^\top/n = \sum_k\hat\lambda_k u^{(k)}(u^{(k)})^\top$ with $(u^{(k)})^\top u^{(l)}=\delta_{kl}$ and $\hat\lambda_1\ge\hat\lambda_2\ge\dots$, and $P_K = \sum_{k\le K}u^{(k)}(u^{(k)})^\top$. The full estimator is $\hat\Theta = \Theta_{\bar\eta} + \Xi X^\top(XX^\top)^{-1}$, where $\Theta_{\bar\eta} = D_{\bar\eta}D_aD_\mu^{-1}(I_r\;\;0_{r\times d})$ is the infinite-sample coefficient on corrupted targets. The post-hoc projected predictor is $\hat y= \hat\Theta P_K x$.

We define the cumulative alignment $A_{ii}(K) = \sum_{k\le K}(u^{(k)}_{i})^2$ and barcode alignment $q_k = \sum_{l>r}(u^{(k)}_l)^2$. Averaging over the fresh test point and the noise $\Xi$, conditional on $X$, gives the per-coordinate prediction MSE
\begin{align}
R(K\mid X)
&\equiv
\frac{1}{r}\EE\left[\left\|y - \hat\Theta P_K x\right\|^2 \mid X\right]
\notag\\
&=
\frac{1}{r}\sum_{j=1}^{r}
\EE \left[
  \left(
    \frac{a_j}{\mu_j}(x - \bar\eta_j P_K x)_j
    - \bigl[\Xi X^\top(XX^\top)^{-1} P_K x\bigr]_j
  \right)^{\!2}
\mid X\right]
\notag\\
&=
  \frac{1}{r}\sum_{j=1}^{r}
  \frac{a_j^2}{\mu_j^2}
  \EE\left[\left(x - \bar\eta_j [P_K x]\right)_j^2\mid X\right]
+
\frac{1}{r}
\EE\left[
  \left\|\Xi X^\top(XX^\top)^{-1} P_K x\right\|^2 \mid X
\right]
\notag\\
&=
\frac{1}{r}\sum_{j=1}^{r}
  \frac{a_j^2}{\mu_j^2}
  \left(
    \mu_j^2
    - 2\bar\eta_j\mu_j^2 A_{jj}(K)
    + \bar\eta_j^2
      \sum_{k,k'\le K} u^{(k)}_j\, u^{(k')}_j\,
      (u^{(k)})^\top\!\Sigma_x\, u^{(k')}
  \right)
  \notag\\
  &\hspace{5cm}
  +
\frac{\tr(\Sigma_\xi)}{rn}
\sum_{k\le K}
  \frac{(u^{(k)})^\top\!\Sigma_x\,u^{(k)}}{\hat\lambda_k}
  \\
&=
\frac{1}{r}\sum_{j=1}^{r}
  \frac{a_j^2}{\mu_j^2}
  \left(
    \mu_j^2
    - 2\bar\eta_j\mu_j^2 A_{jj}(K)
    + \bar\eta_j^2
      \left\{
      \sum_{l=1}^{r} [P_K]_{jl}^2\,\mu_l^2
        +
        \sum_{l=r+1}^{r+d} [P_K]_{jl}^2
      \right\}
  \right)
  \notag\\
  &\hspace{5cm}
  +
\frac{\tr(\Sigma_\xi)}{rn}
\sum_{k\le K}
  \frac{(u^{(k)})^\top\Sigma_xu^{(k)}}{\hat\lambda_k}
\end{align}
where $\Sigma_x = \diag(\mu_1^2,\ldots,\mu_r^2,1,\ldots,1)$ is the population feature covariance. The cross-term vanishes since $\EE[\Xi\mid X]=0$. The variance term uses $P_K(XX^\top)^{-1} = n^{-1}\sum_{k\le K}\hat\lambda_k^{-1}u^{(k)}(u^{(k)})^\top$.

The first term contains off-diagonal overlaps $(P_K)_{lj}$ with $l\neq j$, which are second-order small when $\Sigma_x$ is diagonal and the empirical eigenvectors concentrate near coordinate axes. Setting these off-diagonal terms to zero and using the idempotence of $P_K$ to replace $\sum_{l>r}(P_K)_{lj}^2$ by $A_{jj}(K)(1-A_{jj}(K))$ gives
\begin{equation}\label{eq:rmixed}
R_{\mathrm{diag}}(K)
\!=\! \frac{1}{r}\sum_{j=1}^{r}\left[
    a_j^2\bigl(1-\bar\eta_j A_{jj}(K)\bigr)^2
    + \frac{a_j^2\bar\eta_j^2}{\mu_j^2}\,
      A_{jj}(K)\bigl(1-A_{jj}(K)\bigr)
  \right]
+ \frac{\tr(\Sigma_\xi)}{rn}
\sum_{k\le K}
  \frac{(u^{(k)})^\top\Sigma_xu^{(k)}}{\hat\lambda_k}.
\end{equation}
The first term in the square brackets is the squared mismatch between the clean target and the partially-projected shrunk predictor. The second is a mixing penalty from eigenvectors that partially align with signal coordinate $j$, amplified by $1/\mu_j^2$ for sub-threshold signals; it peaks at $A_{jj}=\tfrac{1}{2}$ and vanishes when each eigenvector is a pure coordinate axis (in which case $A_{jj}=1$). The final term outside the square brackets is the noise variance retained by the projection.

An interior optimal rank requires heterogeneity or mixing. In the ideal separated limit, the retained eigenvectors are pure signal or pure barcode directions, so the mixing penalty vanishes and the optimal rule is to keep the signal directions and discard the barcode directions. Below the separation threshold, increasing $K$ simultaneously recovers signal and admits noise; the optimum balances the two.

\section{Single-Sample Experiments}\label{app:single_sample}
\subsection{Model accuracy in single-sample projection experiments}

\begin{table*}[htbp]
\centering
\caption{Bulk accuracy after sample projection. All accuracies are
percentages. BatchNorm is reset after each projection and recalibrated
by cumulative averaging on equal batches of 250 images. The $k=0$
column is the matched unchanged-weight BatchNorm control.}
\label{tab:single_fisher_eigvec_cumulative_bn_v1}
\begingroup
\scriptsize
\vspace{2mm}
\setlength{\tabcolsep}{2pt}
\renewcommand{\arraystretch}{1.15}

\begin{tabular}{@{}lccccc@{}c@{}lccccc@{}}
\multicolumn{6}{c}{\textbf{Gaussian model}}
& \hspace{1em} &
\multicolumn{6}{c}{\textbf{Noisy-label model (10\% noise)}} \\[2pt]

\cline{1-6}\cline{8-13}
& $k\!=\!0$ & $k\!=\!1$ & $k\!=\!2$ & $k\!=\!4$ & $k\!=\!9$
&&
& $k\!=\!0$ & $k\!=\!1$ & $k\!=\!2$ & $k\!=\!4$ & $k\!=\!9$ \\

\multicolumn{6}{l}{\textit{Fisher source CIFAR-10}}
&&
\multicolumn{6}{l}{\textit{Fisher source uncorrupted}} \\
\cline{1-6}\cline{8-13}

CIFAR-10 train (5k samples)
& 100.0 & 99.9 & 99.9 & 99.9 & 99.8
&&
CIFAR-10 train (5k samples)
& 100.0 & 100.0 & 100.0 & 99.9 & 99.9 \\

CIFAR-10 test
& 84.3 & 83.9 & 83.7 & 83.4 & 83.1
&&
CIFAR-10 test
& 73.4 & 73.1 & 73.0 & 72.7 & 72.3 \\

Gaussian set
& 100.0 & 100.0 & 100.0 & 100.0 & 99.9
&&
Corrupted CIFAR (assigned labels)
& 100.0 & 100.0 & 100.0 & 99.9 & 99.8 \\

\multicolumn{6}{l}{}
&&
Corrupted CIFAR (true labels)
& 0.0 & 0.0 & 0.0 & 0.0 & 0.1 \\

Same-class CIFAR (1k samples)
& 100.0 & 99.3 & 99.1 & 98.9 & 98.7
&&
Same-class CIFAR (1k samples)
& 100.0 & 99.9 & 99.8 & 99.5 & 99.1 \\[+0.5ex]

\multicolumn{6}{l}{\textit{Fisher source Gaussian}}
&&
\multicolumn{6}{l}{\textit{Fisher source corrupted}} \\
\cline{1-6}\cline{8-13}

CIFAR-10 train (5k samples)
& 100.0 & 100.0 & 100.0 & 100.0 & 100.0
&&
CIFAR-10 train (5k samples)
& 100.0 & 100.0 & 100.0 & 100.0 & 99.9 \\

CIFAR-10 test
& 84.3 & 84.0 & 83.9 & 83.8 & 83.4
&&
CIFAR-10 test
& 73.4 & 73.2 & 73.0 & 72.8 & 72.4 \\

Gaussian set
& 100.0 & 98.4 & 96.1 & 90.5 & 75.6
&&
Corrupted CIFAR (assigned labels)
& 100.0 & 100.0 & 100.0 & 99.9 & 99.8 \\

\multicolumn{6}{l}{}
&&
Corrupted CIFAR (true labels)
& 0.0 & 0.0 & 0.0 & 0.1 & 0.1 \\

Assigned-label class (1k samples)
& 100.0 & 100.0 & 100.0 & 100.0 & 99.9
&&
Assigned-class CIFAR (1k samples)
& 100.0 & 99.9 & 99.9 & 99.8 & 99.7 \\

Source-class CIFAR (1k samples)
& 100.0 & 100.0 & 100.0 & 100.0 & 100.0
&&
True-class CIFAR (1k samples)
& 100.0 & 100.0 & 100.0 & 100.0 & 99.8 \\
\cline{1-6}\cline{8-13}
\end{tabular}

\endgroup
\end{table*}

\clearpage

\end{document}